# Incremental Learning Framework Using Cloud Computing

Kumarjit Pathak [a*] , Prabhukiran G [b*], Jitin Kapila [c*], Nikit Gawande [d]

[a] Data Scientist professional , Harman , Whitefield, Bangalore ,mail:Kumarjit.pathak@outlook.com

[b] Data Scientist professional , Harman , Whitefield, Bangalore ,mail:Prabhukiran.g@gmail.com

[c] Data Scientist professional , Zeta Global , Indiranagar, Bangalore ,mail:Jitin.kapila@outlook.com

[b] Data Scientist professional , Northwestern University , ,mail: nikitgawande2018@u.northwestern.edu

**Abstract:** High volume of data, perceived as either challenge or opportunity. Deep learning architecture demands high volume of data to effectively back propagate and train the weights without bias. At the same time, large volume of data demands higher capacity of the machine where it could be executed seamlessly.

Budding data scientist along with many research professionals face frequent disconnection issue with cloud computing framework (working without dedicated connection) due to free subscription to the platform. Similar issues also visible while working on local computer where computer may run out of resource or power sometimes and researcher has to start training the models all over again.

In this paper, we intend to provide a way to resolve this issue and progressively training the neural network even after having frequent disconnection or resource outage without loosing much of the progress.

## INTRODUCTION

Most of the algorithm follows some variant of gradient descent optimizer to find effective global minima where the cost is least with respect to the value of the tuning parameter and weights. Gradient descent uses the loss function to minimize over each iteration. We can imagine loss as the difference between what we observe in the training data and the prediction. For an example if we are predicting sales then loss = actual sales – predicted sales for each observation. For any classification problem, we can imagine loss as the difference between predicted probability and actual class. There

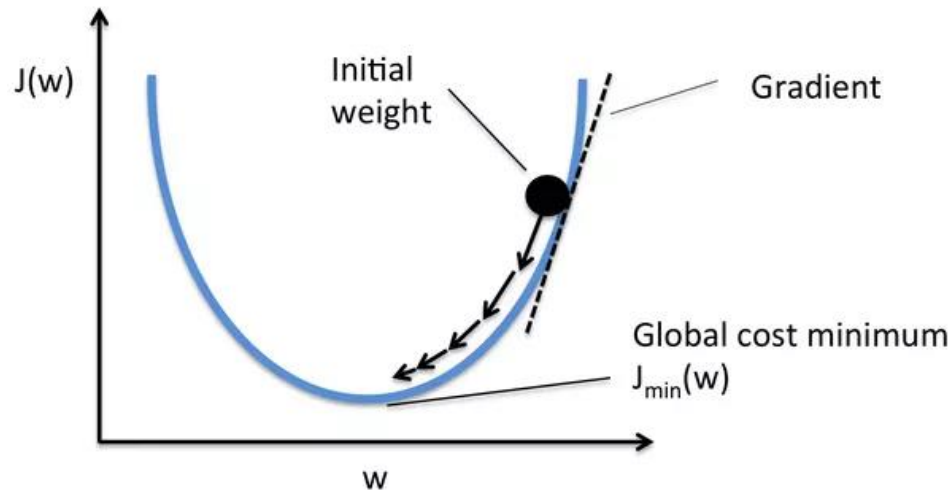

Fig. 1. Gradient descent algorithm et el [1].

are different formulas to calculate loss

- Cross-Entropy Loss
- Hinge Loss
- Huber Loss
- Kullback-Leibler Loss
- L1 Loss
- L2 loss
- Maximum Likelihood Loss
- Mean Squared Error

And many more. Learning mechanism is to start our algorithm to learn on the training data update the weights 'W' of the algorithm to gradually minimize the loss 'J(W)' at each iteration. This process goes on until the convergence i.e algorithm reaches optimum value of 'W' for which 'J(W)' is minimum for the training sample. Above graph is just to provide intuition to this theory.

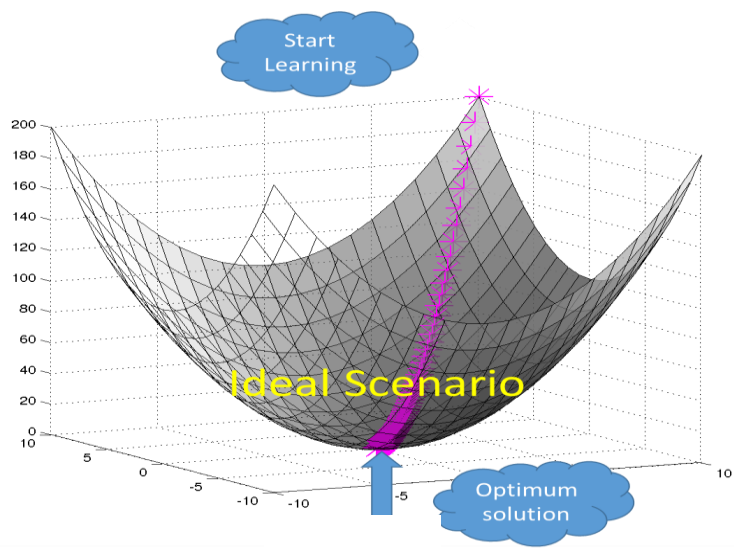

Fig. 2.  Ideal training scenario

In an Ideal situation, we all expect to have a happy path from where the algorithm started until the point of optimization. In practice, this may not be the case due to service interruption of network or the cloud computing systems.

This is not simple with the complexity of the problem we deal with regularly and the amount of data we use for training. Deep learning model takes days of continuous training until it optimizes.

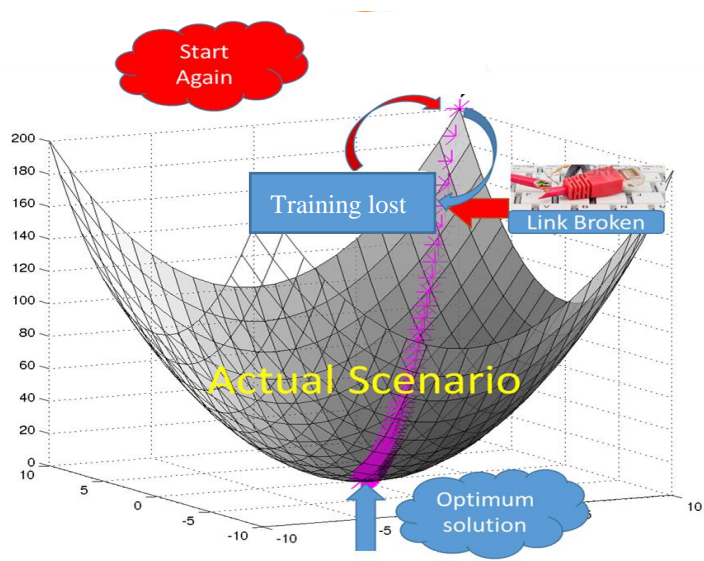

Fig. 3.  Actual training scenario

More appropriate for the researchers who are using free access provided by Google, AWS (with free account only) etc. If by any chance the connection terminates we do not get back the same instance again and loss on valuable time spend on the training. We need to start the training again.

We might have COLAB account to use cloud computing and every 12 hours the instance automatically is terminated. In addition, similar situation we see in regular internal server setup as well or rather say personal computer where power-saver not turned off.

With the approach proposed in this paper, we can avoid training from the scratch every time we have a connection issue or any other disconnection. We can progressively train algorithm, without loosing any significant important recent weight update.

## I.  LITERATURE SURVEY

With development of deep learning techniques and new advancement in the technology more and more computing capacity is on the demand. Our work is mainly inspired by our experience in having consistant disconnection to the cloud computing platform like "colab", "AWS" where again and again we have lost the valuable training time and was unable to recover the same instance. While reviewing articles on the internet we found a good review paper *"Incremental learning algorithms and applications"* by Alexander Gepperth, Barbara Hammer published on European Symposium on Artificial Neural Networks, Computational Intelligence and Machine Learning et el [2]. Authors have mainly focused on live stream data adaptation to avoid concept drift over time. Covariate shift over time is a genuine issue in data science practice however there is a dilemma for how much of the new trend of the data to be adapted. There can be situation where the new changes in the data may be a temporary issue or even if it is a real market scenario how much the algorithm to be tuned to the recent changes. Learning new trend is directly proportional to forgetting old dynamics of the data thus, gives rise to *stability-plasticity dilemma*. Author has suggested explicit partitioning approach, prototype based method and ensembling of models to reduce such issues.

Ronald Kemker and Christopher Kanan , in the paper " [8] "FEARNET:  BRAIN-INSPIRED  MODEL  FOR INCREMENTAL LEARNING", has addressed catastrophic forgetting of ANN's with a brain like design of the network to create a memory reply employs a generative auto-encoder for pseudorehearsal, which mitigates catastrophic forgetting by generating previously learned examples that are replayed alongside novel information during consolidation. This process does not involve storing previous training data as stated by author.

Ghouthi Boukli Hacene, Vincent Gripon, Nicolas Farrugia, Matthieu Arzel & Michel Jezequel et el[9] has described data augmentation approach using nearest class mean (NCM). Methond combines DNN and transfer learning along with majority voting achieves incremental training accuracy on the image classification data.

**Input:** For each database drawn from $\mathcal{D}_k$, $k=1,2,....,K$
- Sequence of $m$ training examples $S=[(x_1,y_1),(x_2,y_2),....,(x_m,y_m)]$.
- Weak learning algorithm WeakLearn.
- Integer $T_k$, specifying the number of iterations.

**Do for** $k=1,2,....,K$:

**Initialize** $w_1(i) = D(i) = 1/m, \forall i$, unless there is prior knowledge to select otherwise.

**Do for** $t = 1,2,....,T_k$:

1.  Set $D_t = w_t / \sum_{i=1}^{m} w_t(i)$ so that $D_t$ is a distribution.
2.  Randomly choose training $TR_t$ and testing $TE_t$ subsets according to $D_t$.
3.  Call WeakLearn, providing it with $TR_t$.
4.  Get back a hypothesis $h_t : X \to Y$, and calculate the error of $h_t$ : $\varepsilon_t = \sum_{i:h_t(x_i)\neq y_i} D_t(i)$ on

    $S_t = TR_t + TE_t$. If $\varepsilon_t > \frac{1}{2}$, set $t = t - 1$, discard $h_t$ and go to step 2. Otherwise, compute normalized error as $\beta_t = \varepsilon_t / (1 - \varepsilon_t)$.
5.  Call weighted majority, obtain the composite hypothesis $H_t = \arg\max_{y \in Y} \sum_{t:h_t(x)=y} \log(1/\beta_t)$,

    and compute the composite error $E_t = \sum_{i:H_t(x_i)\neq y_i} D_t(i) = \sum_{i=1}^{m} D_t(i)[|H_t(x_i) \neq y_i|]$

    If $E_t > \frac{1}{2}$, set $t = t - 1$, discard $H_t$ and go to step 2.
6.  Set $B_t = E_t/(1-E_t)$ (normalized composite error), and update the weights of the instances:

    $w_{t+1}(i) = w_t(i) \times \begin{cases} B_t, & if \ H_t(x_i) = y_i \\ 1, & otherwise \end{cases}$

    $= w_t(i) \times B_t^{1-[|H_t(x_i)\neq y_i|]}$

**Call weighted majority** on combined hypotheses $H_t$ and **Output** the final hypothesis:

$$H_{final} = \arg\max_{y \in Y} \sum_{k=1}^{K} \sum_{t:H_t(x)=y} \log\frac{1}{B_t}$$

Fig. 4.  Learn++: An Incremental Learning Algorithm for Supervised Neural Networks. Et el [7]

Robi Polikar, Lalita Udpa, Satish S. Udpa, Vasant Honavar et el[7] has experimented on Adaboost's distribution update rule. Author uses weak learners to build a rough estimate of the decision boundary thus enabling faster training. These weak learners learns multiple decision boundary using different subset of the data and hence segmented learning achieved on each set of week learners. This is further ensambled with weighted voting to generate the classification.

An example of the algorithm as stated by the authors depicted in fig.4.

Leon  Bottou  et  el  [3]  has

showcased the effectiveness of SGD( Stochastic Gradient Descent) to effectively achieve incremental learning through a consistent data set where entire data is memory intensive and gradual parameter update is the way to go.

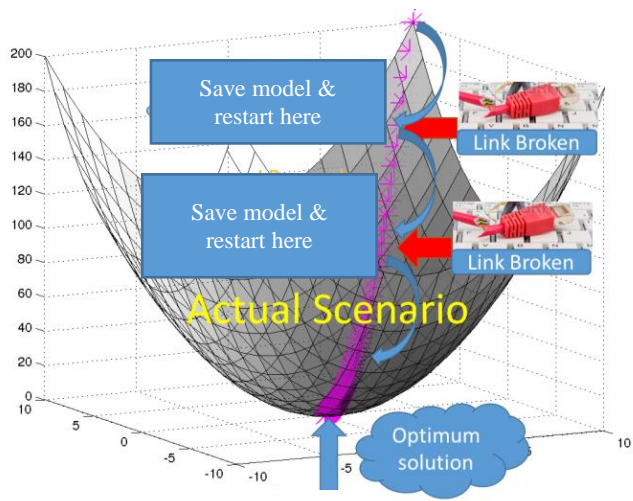

Fig. 5.  Solution to frequent disconnection.

## II.  METHOD AND PROCEDURE

Our approach address the continuous network disconnection issue and provides an way that learning can be saved at regular interval and while any disconnection we would have the latest copy of the learned parameters.

This approach is rather inspired by the Stochastic Gradient Descent which enables a model to incrementally learn the data dynamics.

Two aspect if we can take care it solves our purpose:

• To track different hyper-parameters at each stage of the learning process at regular interval.

• Continuously overwrite the most updated weights at regular interval

If we can take care of this simple staff then we can use previously saved model along with all the hyper-parameters as a base- model and incrementally learn from the same.

In case if researcher is using colab or AWS, researcher can create a local folder in the local computer and as the training goes on incrementally save the best models using some condition on loss decrement or accuracy improvement for continuously few epochs.

```
!pip install -U -q PyDrive
from pydrive.auth import GoogleAuth
from pydrive.drive import GoogleDrive
from google.colab import auth
from oauth2client.client import GoogleCredentials

auth.authenticate_user()
gauth = GoogleAuth()
gauth.credentials = GoogleCredentials.get_application_default()
drive = GoogleDrive(gauth)
```

```
# Create GoogleDriveFile instance with title '.
file2 = drive.CreateFile({'title': 'kp'})    # note this will create a new file on google drive
file2.Upload() # upload the file.
print('title: %s, id: %s' % (file2['title'], file2['id']))
```

```
file2 = drive.CreateFile({'id':'1xtKWAHQ1gf6Lw 9wrAXXXXX'})
file2.GetContentFile('kp')
```

```
file = open('kp', 'w')
```

Fig. 6.  File creation for saving the model in google drive directly from COLAB.

We have experimented using colab where we have followed the following steps which has worked really well and even after any disconnection we have successfully avoided any large loss in training time.

**Step1:** *Provide google drive authentication for colab.*

**Step2:** *Create a file in the google drive*

**Step3:** *Open the file in write mode.*

**Step4:** *Start training the model and periodically over write the weights on the google drive file which is currently open.*

We have used Keras function "model.save", which provides with facility to save the architecture of the model, allowing to re-create the model, weights of the model, training configuration (loss, optimizer), state of the optimizer, allowing to resume training exactly where you left off.

Implementation of the same architecture can be found here.

III.   CONCLUSION

We have highlighted an approach, which essentially avoids training time loss due to frequent disconnection with cloud platform. Effectiveness of this approach is justified with experiments. This would help researchers and data scientists without having much computation power in the local computer to effectively use cloud computing platform to train the models without thinking of frequent disconnection of network.